\documentclass[conference]{IEEEtran}
\IEEEoverridecommandlockouts
\usepackage{cite}
\usepackage{amsmath,amssymb,amsfonts}
\usepackage{algorithmic}
\usepackage{graphicx}
\usepackage{textcomp}
\usepackage{xcolor}
\usepackage{float}
\def\BibTeX{{\rm B\kern-.05em{\sc i\kern-.025em b}\kern-.08em
    T\kern-.1667em\lower.7ex\hbox{E}\kern-.125emX}}
\usepackage{url}
\begin{document}

\title{Introducing AI-Driven IoT Energy Management Framework\\

}

\author{\IEEEauthorblockN{Shivani Mruthyunjaya}
\IEEEauthorblockA{\textit{Texas State University} \\
San Marcos, USA \\
aon9@txstate.edu}
\and
\IEEEauthorblockN{Anandi Dutta}
\IEEEauthorblockA{\textit{Texas State University} \\
San Marcos, USA \\
anandi.dutta@txstate.edu}
\and
\IEEEauthorblockN{Kazi Sifatul Islam}
\IEEEauthorblockA{\textit{Texas State University} \\
San Marcos, USA \\
kazi\_sifat@txstate.edu}
}

\maketitle

\begin{abstract}
Power consumption has become a critical aspect of modern life due to the consistent reliance on technological advancements. Reducing power consumption or following power usage predictions can lead to lower monthly costs and improved electrical reliability. The proposal of a holistic framework to establish a foundation for IoT systems with a focus on contextual decision making, proactive adaptation, and scalable structure. A structured process for IoT systems with accuracy and interconnected development would support reducing power consumption and support grid stability. This study presents the feasibility of this proposal through the application of each aspect of the framework. This system would have long term forecasting, short term forecasting, anomaly detection, and consideration of qualitative data with any energy management decisions taken. Performance was evaluated on Power Consumption Time Series data to display the direct application of the framework.  
\end{abstract}


\section{Introduction}
 The power grid plays a critical role in ensuring a reliable and efficient energy supply, adapting to fluctuating demand while maintaining stability. Traditionally, grid operations relied on manual monitoring, scheduled maintenance, and reactive decision-making, where human operators adjusted the power distribution based on historical trends and field reports. However, due to the over-reliance on technology, demand on the power grid has surged, requiring more adaptive and proactive management strategies. Today, the incorporation of IoT systems is transforming the power grid into a smart, self-regulating network, capable of real-time monitoring, predictive maintenance, and optimized energy distribution. In this study, we proposed several measures, such as long-term forecasting, short-term forecasting, and anomaly detection, for a holistic energy usage management framework. We applied several deep learning algorithms and machine learning algorithms in that proposed framework. 

\section{System Model and Problem Formulation}

IoT systems have been successfully implemented in smart homes, capable of recognizing homeowner schedules and dynamically adjusting power consumption to minimize energy waste. This not only reduces household energy costs, but also contributes to a more balanced and efficient grid by improving power forecasting. Several studies have explored IoT-driven energy management, each taking different approaches to forecasting power consumption. One study proposed an ensemble learning approach leveraging Decision Trees (DT), Random Forests (RF), and eXtreme Gradient Boosting (XGBoost) for long-term power consumption forecasting [1]. Another study applied a neural network-based model, demonstrating the effectiveness of deep learning techniques in capturing complex energy consumption patterns [2] . These studies highlight the importance of both long-term and short-term forecasting in energy management, as different forecasting methods support different grid-level decisions.
While accurate forecasting is crucial for optimizing power distribution, another critical aspect of grid reliability is to proactively respond to anomalies and critical events. Current IoT systems, alert users after irregularities occur, rather than preventing them. In addition, when blackouts occur IoT systems may be able to ensure the demand is fulfilled by creating a dependency on energy produced by renewable sources. To address this limitation and support households, this paper proposes a holistic IoT framework that integrates both long-term and short-term forecasting models, anomaly detection mechanisms, and contextual qualitative factors. By bridging the gap between predictive analytics and real-time adaptation, the proposed system aims to enhance grid reliability, energy efficiency, and proactive response mechanisms, ultimately creating a more resilient and intelligent power infrastructure.

\section{Methodology}

The holistic framework proposed capitalizes on the connective nature of IoT systems, while considering each aspect affecting the power grid. The power consumption by households has been the main focus in IoT systems to focus on the energy management without attention placed on proactive effort towards sustainability.

  The framework (Figure 1) consists of 4 different components: long-term forecasting, short-term forecasting, anomaly detection, and contextual integration, each contributing to improved energy optimization and power grid stability.

\subsection{ Long-Term Forecasting} 
Long-term forecasting holds a critical role in optimizing energy management and infrastructure planning within power grids. Accurate long-term predictions enable power grids to implement strategic changes with confidence, mitigating risks associated with costly modifications and large-scale operational adjustments. Effective long-term forecasting not only facilitates informed decision-making but also provides IoT systems with a historical reference for energy consumption patterns, enhancing their adaptability and long-term viability. Furthermore, numerous studies have demonstrated that neural networks are among the most effective models for power consumption forecasting, as they excel in identifying complex, nonlinear patterns inherent in human energy usage behavior. Several studies support the use of Long Short-Term Memory (LSTM) networks for power consumption forecasting, given their effectiveness in capturing temporal dependencies in time-series data [2]. While ARIMA [5] is a common approach for time series forecasting, it struggles with capturing long-term dependencies, making LSTM a more suitable choice. By integrating long-term forecasting, IoT systems can adapt to changing power consumption trends and provide a historical understanding of energy behavior. 

 \begin{figure}[H]
    \centering
    \includegraphics[width=9cm]{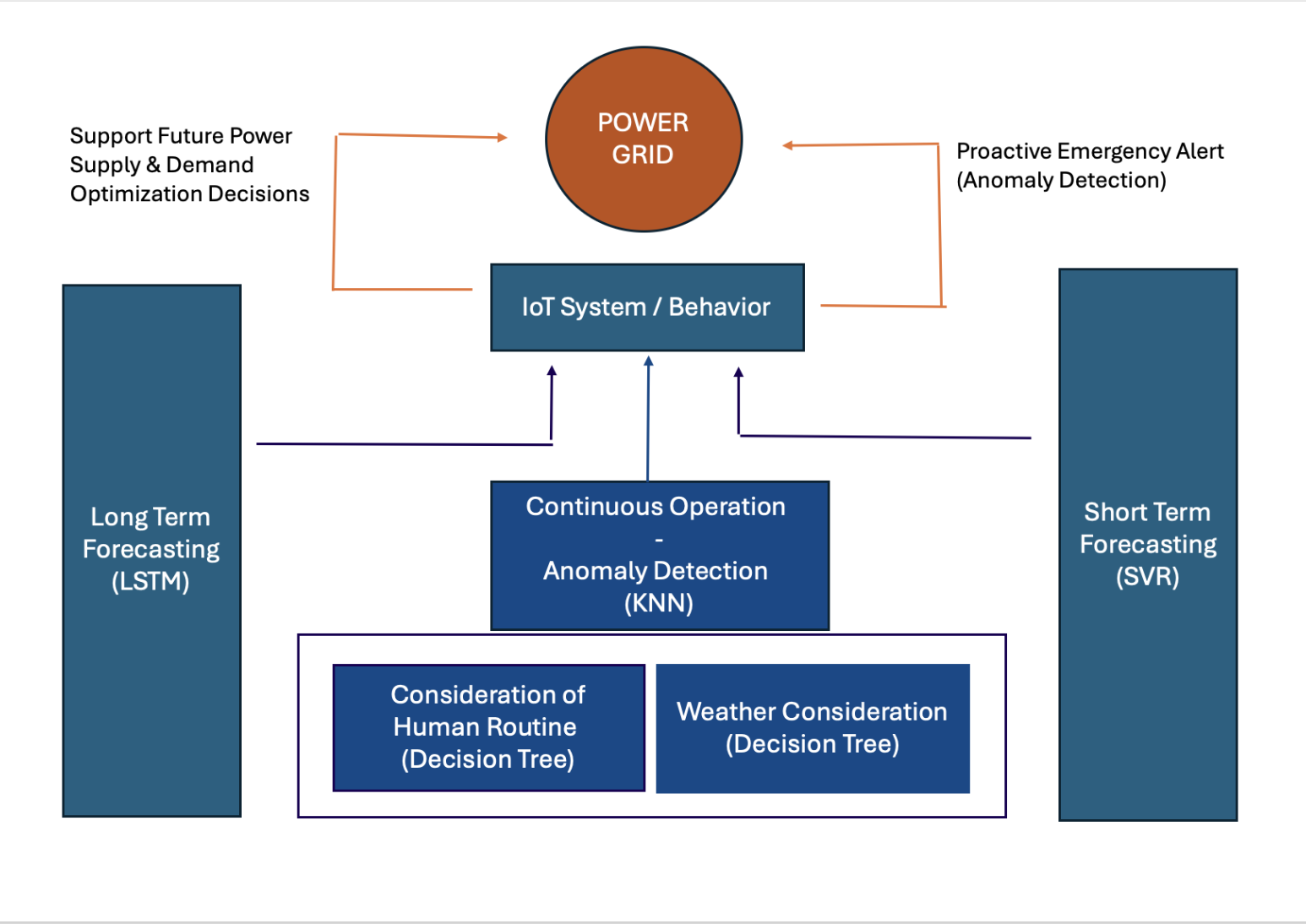}
    \caption{Proposed IoT System Framework}
    \label{Task 1: Patient Centric Report Generation Visualization}
\end{figure}
  
\subsection{ Short-Term Forecasting} 
Short-term analysis is critical for real-time grid management, dynamic energy distribution, and the prevention of power failures. Unlike long-term forecasting, which supports strategic planning, short-term forecasting ensures that power grids maintain a stable balance between supply and demand. This allows grid operators to dynamically adjust power, preventing shortages, overloads, and potential blackouts through areas using IoT systems. In respect to utility companies, predicting immediate consumption fluctuations, utility companies can reduce reliance on expensive emergency power generation, minimize energy wastage, and enhance demand-response programs based on short-term trends. This proactive approach ensures a more efficient allocation of resources and prevents unnecessary operational costs.
	While neural networks are highly effective, they are computationally expensive and may overfit small datasets, making them unsuitable for frequent computations required in short-term analysis. Instead, Support Vector Regression (SVR) is utilized for short-term forecasting due to its efficiency and ability to handle non-linear patterns [9]. 
Another important application of short-term forecasting is the enhancement of anomaly detection. Short-term forecasting involves evaluating whether real-time consumption behavior aligns with both long-term projections and prior short-term predictions. This adaptive approach enables IoT systems to detect and adapt to gradual shifts in household energy usage. On a larger scale, if widespread deviations from predicted patterns occur across multiple households, a decision tree model can assess the variations and alert the grid to potential anomalies, such as sudden weather changes or imminent power surges.
\subsection{ Anomaly Detection}
Anomaly detection is a critical aspect of the proposed framework, ensuring both household-level and grid-wide reliability. Unexpected power consumption patterns may indicate potential equipment failures, power surges, or external disruptions. This framework also emphasizes anomaly detection, which plays a crucial role in identifying irregularities in power consumption. Significant deviations in energy usage across multiple households may indicate potential equipment malfunctions, impending power surges, or other critical grid disruptions. On an individual level, anomaly detection enables households to receive alerts regarding unusual consumption patterns, allowing them to assess and address potential issues proactively. Anomalies can be detected by analyzing whether current power consumption significantly deviates from historical data or expected values. By integrating anomaly detection into the framework, the system can enhance grid reliability, minimize disruptions, and improve energy efficiency through proactive intervention. The framework employs k-Nearest Neighbors (k-NN) for anomaly detection [10]. Additionally, Decision Tree Classifiers are proposed to integrate contextual information, assessing factors such as temperature, weather conditions, time of day, and household behavior to refine anomaly detection accuracy.
\subsection{Contextual Integration}
While the proposed framework primarily focuses on quantitative analysis, the consideration of qualitative factors is also essential for enhancing real-time decision-making within IoT systems. While historical consumption patterns and predictive models can provide valuable insights, they may not fully capture the dynamic nature of real-world energy usage. To address this, the framework suggests integrating a Decision Tree model for anomaly detection. This model would assess deviations in power consumption by comparing a given data point with historical records and short-term forecasts but also incorporate contextual variables. Factors such as current weather conditions, temperature, time of day, day of the week, holidays, and the consumption behavior of other IoT-enabled households play a crucial role in distinguishing between anomalies and significant shifts in energy demand that require adaptation. Weather conditions and temperature directly influence power consumption patterns, as extreme temperatures lead to increased heating or cooling demands. Similarly, time-based factors such as the time of day and the day of the week contribute to variations in energy usage, with peak consumption typically occurring during specific hours when household and industrial activities are at their highest.
Holidays introduce additional energy fluctuations, as residential energy consumption may rise while commercial  energy consumption may decline due to closures. The behavior of other IoT-enabled households further provides valuable contextual information, allowing for a collective assessment of energy trends. If a large deviation in power consumption is detected across multiple households, it may indicate a broader trend, such as a regional shift in consumption habits, rather than an isolated anomaly.
By integrating these qualitative and contextual factors into the anomaly detection framework, the system can differentiate between unexpected irregularities and meaningful changes in energy demand [11]. This ensures that power grids can respond appropriately—whether by addressing potential faults and inefficiencies or adapting to new consumption behaviors to enhance energy distribution and grid stability.
While quantitative analysis provides essential insights, qualitative factors significantly impact energy consumption behavior. Although Decision Trees were not implemented in this study, they are theoretically suited for integrating these factors, as they can process categorical and numerical data efficiently.
\subsection{ Model Implementation}
To establish the feasibility of the proposed framework, the following models were implemented:
Long-Term Forecasting: LSTM Neural Network
Short-Term Forecasting: Support Vector Regression (SVR)
Anomaly Detection: k-Nearest Neighbors (k-NN)
Contextual Integration: Theoretical Decision Tree Classifier
Through integrating these methodologies, the framework aims to enhance power grid stability through proactive and adaptive energy management, ensuring a scalable and computationally efficient approach to energy management. 
\subsection{ Dataset Discussion}
The dataset utilized in this study was sourced from Kaggle and comprises power consumption data collected from a single household at one-minute intervals over a six-month period: from January 2007 to June 2007. This dataset was selected due to its high-resolution temporal granularity, making it well-suited for analyzing power consumption patterns, forecasting energy demand, and detecting anomalies. The dataset consists of nine key features and includes a total of 260,640 measurements, providing a comprehensive view of energy consumption behavior. While the dataset provides high-resolution energy consumption data, real-world IoT applications must also address storage constraints, processing efficiency, and scalability when handling large-scale power grid data [12].
Each data entry in the dataset includes date and time stamps, which allow for precise chronological analysis of energy usage trends. The primary consumption metric, Global Active Power (kW), represents the total power consumed by the household, while Global Reactive Power (kW) accounts for the energy that does no useful work but is required for maintaining voltage levels in inductive loads. Additionally, the dataset includes Voltage (V), which provides insights into fluctuations in household electricity supply, and Global Intensity (A), which represents the total current flowing through the circuits. To further segment energy consumption, the dataset contains three sub-metering features: Sub-metering 1 records power usage from kitchen appliances, Sub-metering 2 captures energy consumption from laundry-related devices, and Sub-metering 3 measures the electricity usage of high-energy appliances such as electric water heaters and air conditioners.
To prepare the dataset for analysis, several pre-processing steps were undertaken. Firstly, the data was structured into a time-series format, ensuring that sequential dependencies were preserved for forecasting applications. Given the importance of standardization in machine learning models, a MinMaxScaler transformation was applied to all numerical features, scaling values within a range of (0,1) to improve model convergence and training efficiency. Notably, outliers were not removed from the dataset, they are required to show a key component of the proposed framework. These outliers could correspond to sudden spikes in energy consumption due to equipment malfunctions or abnormal household behavior, which are valuable for refining anomaly detection models.
Following pre-processing, the dataset was divided into training and testing subsets, with 80\% of the data allocated for training and 20\% reserved for testing. No separate validation set was created, as the primary focus of this study was on assessing real-time energy forecasting and anomaly detection performance rather than fine-tuning hyperparameters extensively. The decision to use an 80-20 split ensures that the models receive sufficient data for learning consumption patterns while still having enough unseen data for evaluating generalization capabilities.
The dataset’s real-world nature makes it highly relevant for developing IoT-integrated energy management systems, particularly in the context of short-term and long-term power consumption forecasting. The dataset’s fine-grained temporal resolution and appliance-level breakdown make it well-suited for smart home IoT energy optimization, as it allows for detailed consumption tracking and pattern identification. The dataset used in this study remains accessible through Kaggle and serves as a foundational component in evaluating the effectiveness of the proposed methodology. 

\subsection{ Process Undertaken For Implementation}
	The implementation of the Long-Short Term Memory (LSTM) network on the selected dataset was made with a focus on ensuring optimal functioning, low error measurements and mitigating overfitting. To ensure data standardization and efficient model convergence, a MinMaxScalar normalization was applied, the actualization of a uniform scale supports the recognition of complex patterns.  
 In regard to the forecasting process, both long term and short, the process of deciding on an optimal “look-back” hyper parameter value is vital to model performance. The “look-back” value defines the sequence length of past data points used to predict a future time period. Initial experimentation with a “look-back” value of 30, while computationally successful, resulted in overfitting. As expected due to this the natural inclination of Long-Short Term Memory (LSTM) networks to overfit, when dealing with smaller datasets. In contrast, a large lookback value of 400 resulted in poor generalization to the dataset, as presented by the substantial validation loss.  Following an iterative process of look-back value identification, a value of  lookback value of 100 was implemented to minimize overfitting, achieve a relatively low loss, and ensure a sufficient amount of data was reviewed for pattern recognition.    Additionally, the batch size chosen for the LSTM training process was optimized to a value of 1024 for maximized GPU utilization, as variation in batch sizes resulted in undesirable error and loss. This decision supported the model’s ability to generalize to new data, and stabilized the stability of the training process. 
 For short term forecasting, the Support Vector Machine (SVM) was implemented with similar decisions and considerations undertaken [7]. The Support Vector Machine (SVM) is established as being less prone to overfitting, regardless of the size of the dataset. Enabling the choice to utilize a smaller lookback value of 30, as it would be large enough to capture the existing short term dependencies. 
 The process for the identification of anomalies utilizing the KNN algorithm within the time series dataset was completed with a specialized process. To ensure that anomalies were detected with respect to the temporally relevant patterns, the dataset was reformatted into sectioned windows. Then calculation regarding the distance of a certain window in reference to its neighbors would be evaluated. Finally, values with a significant variation in distance to neighbors, determined by its percentile in regard to other distances, would be identified and visualized for analysis.  

\section{Results}
    Implementation of the LSTM upon the discussed dataset reinforced the expected assumption regarding both accuracy and the ability to perform such aspects of forecasting. The comparison is supported by the relatively low error measurements for both training and testing. 

\begin{align}
    \begin{tabular}{ c c}
       Error Name: & Value \\ 
     \hline
      Train Mean Absolute Error & 0.121094 \\ 
     \hline
      Train Root Mean Squared Error & 0.3040671\\ 
     \hline
      Test Mean Absolute Error &  0.109812\\  
      \hline
      Test Root Mean Squared Error & 0.292612 \\
      \hline
    \end{tabular}
\end{align}

Table 1 : Neural Network Long Term Forecasting: Error Measurement For Training and Test Data	

\vspace{5mm} %5mm vertical space
The Long Short Term Memory (LSTM) model demonstrates moderate predictive performance, as supported by the error metrics. As presented, on the training set of the dataset, the model yielded a Mean Absolute Error (MAE) of 0.1211 and a Root Mean Squared Error (RMSE) of 0.3041. On the test section of the dataset, better results with a Mean Absolute Error (MAE) of 0.109812 and a Root Mean Squared Error (RMSE) of 0.292612. The metrics presented were computed on a dataset normalized through the application of the MinMaxScaler, therefore indicating that the model’s prediction deviates from the actual power consumption by 10.98\% to 12.11\% of the normalized range.  

The model demonstrates marginally better performance on the test set compared to the training set, as displayed by the lower MAE and RMSE values. This suggests that the model is not over-fitting, a common concern with LSTM-forecasting with time series data. In contrast, the higher RMSE values compared to MSE values, for both testing and training, convey the existence of some prediction errors. Prediction errors in this manner are a characteristic of LSTM model application for minimal time series data, but can be mitigated through larger quantities of time series data. 

The overall observation of the error metrics indicate that the model was strongly capable of capturing the temporally relevant patterns of the time series.

\begin{figure}[H]
    \centering
    \includegraphics[width=10cm]{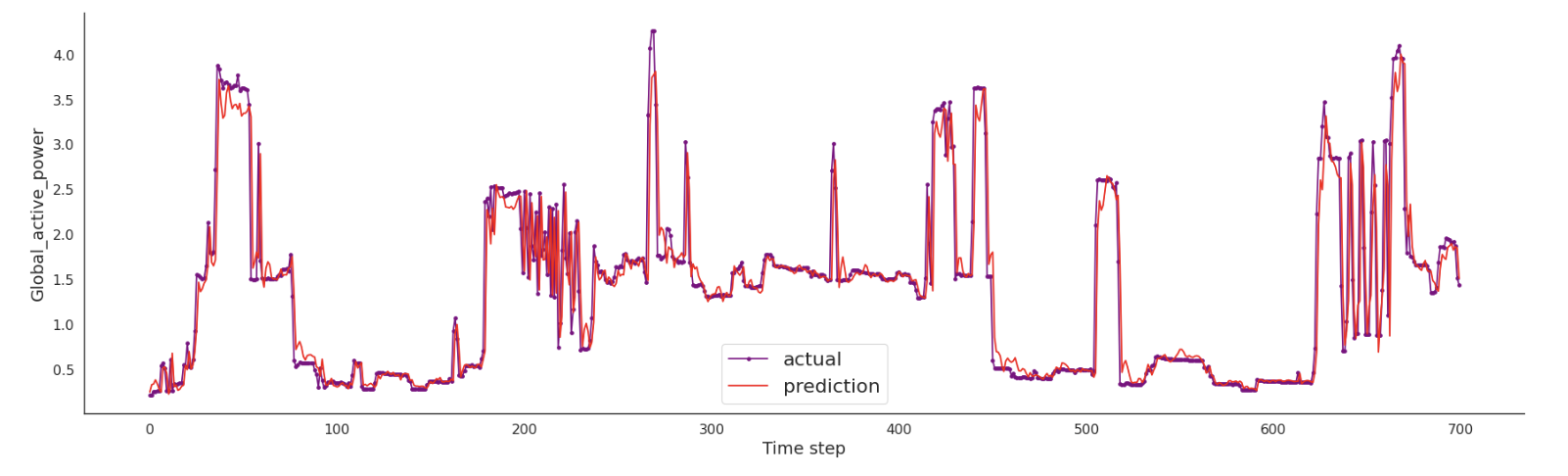}
    \caption{Visualization of Long Term Forecasting Through LSTM}
    \label{fig:Visualization of Long Term Forecasting Through LSTM}
\end{figure}

Visualization of the forecasting was conducted to support the notion that the LSTM [6] would be able to accurately predict power consumption through its natural ability to recognize complex patterns within usage. The prediction of the power consumption follows the actual consumption very closely, while there do seem to be some situations in which the actual consumption seems to be over the predicted. This may cause some issues, the acknowledgment of this discrepancy, would allow the grid and the IoT system to implement a specific offset amount to ensure that the predicted value is reasonably higher to ensure that individuals are supported in their power consumption behavior.

 \begin{figure}[H]
    \centering
    \includegraphics[width=9cm]{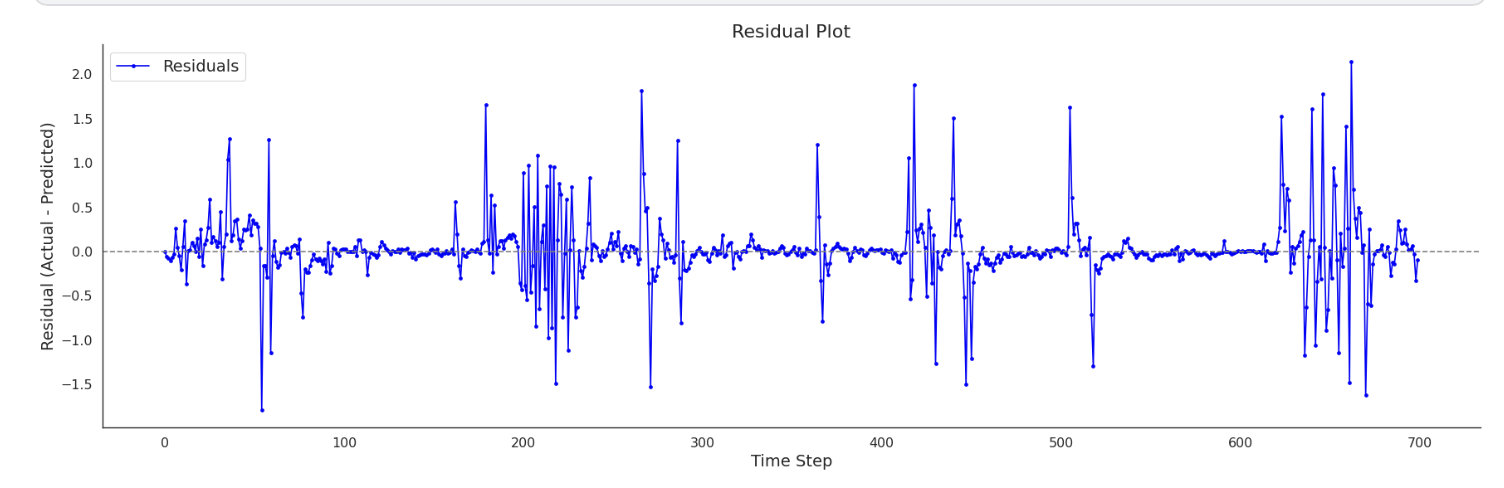}
    \caption{Residual of Long Term Forecasting Through LSTM}
    \label{fig:Residual of Long Term Forecasting Through LSTM}
\end{figure}

The direct observation of residual visualization further represents that their is a certain extent of difference between the actual and expected power consumption, these differences may be attributed to external variables that the model may not be able to fully take into consideration such as weather, changes in human routine during that time period the test dataset presented. Further supporting the importance of ensuring that the power supplied allows for fluctuation and variability. 

Enabling IoT systems to perform long-term analysis assists their standardization and sustainability, ensuring a foundation for future development through the implementation of neural network-based forecasting.
Next, to display the ability of short term analysis to be implemented within the IoT system, Short Term Forecasting was implemented on the discussed dataset. The accuracy in regards to prediction was as expected. The absolute error amount seems to be significantly smaller than that achieved by the implemented LSTM. The accuracy in forecasting displays the ability of the SVM to be implemented for the previously presented endeavor. While also supporting its implementation in long term forecasting [8], the LSTM would be better suited to identify complexities and signal for long term changes to be implemented with greater confidence than an SVM.  

\begin{align}
    \begin{tabular}{ c c}
       Error Name: & Value \\ 
       \hline
      Test Mean Absolute Error &  0.03594\\  
      \hline
      Test Root Mean Squared Error & 0.04166 \\
      \hline
    \end{tabular}
\end{align}

Table 2. SVM Short Forecasting: Error Measurement For Training and Test Data	

\vspace{5mm} %5mm vertical space

The performance of the Support Vector Machine (SVM) model, as indicated by a Test Mean Absolute Error (MAE) of 0.03594 and a Test Root Mean Error (RMSE) of 0.04166, suggests strong predictive accuracy. The MAE of approximately 0.036 implies that on average, the model’s forecasts deviate from the actual power consumption by about 3.6\% of the normalized range. The low MAE metric presents that the model maintained a high degree of precision across forecasting in the test dataset. 
The RMSE value of 0.04166, which exceeds the MAE slightly, reinforces the accuracy and precision of the model during short term forecasting. The RMSE is only approximately 16\% higher than the MAE, indicating that the model maintains a relatively high level of precision and is not heavily impacted by large individual errors.

\begin{figure}[H]
    \centering
    \includegraphics[width=7cm]{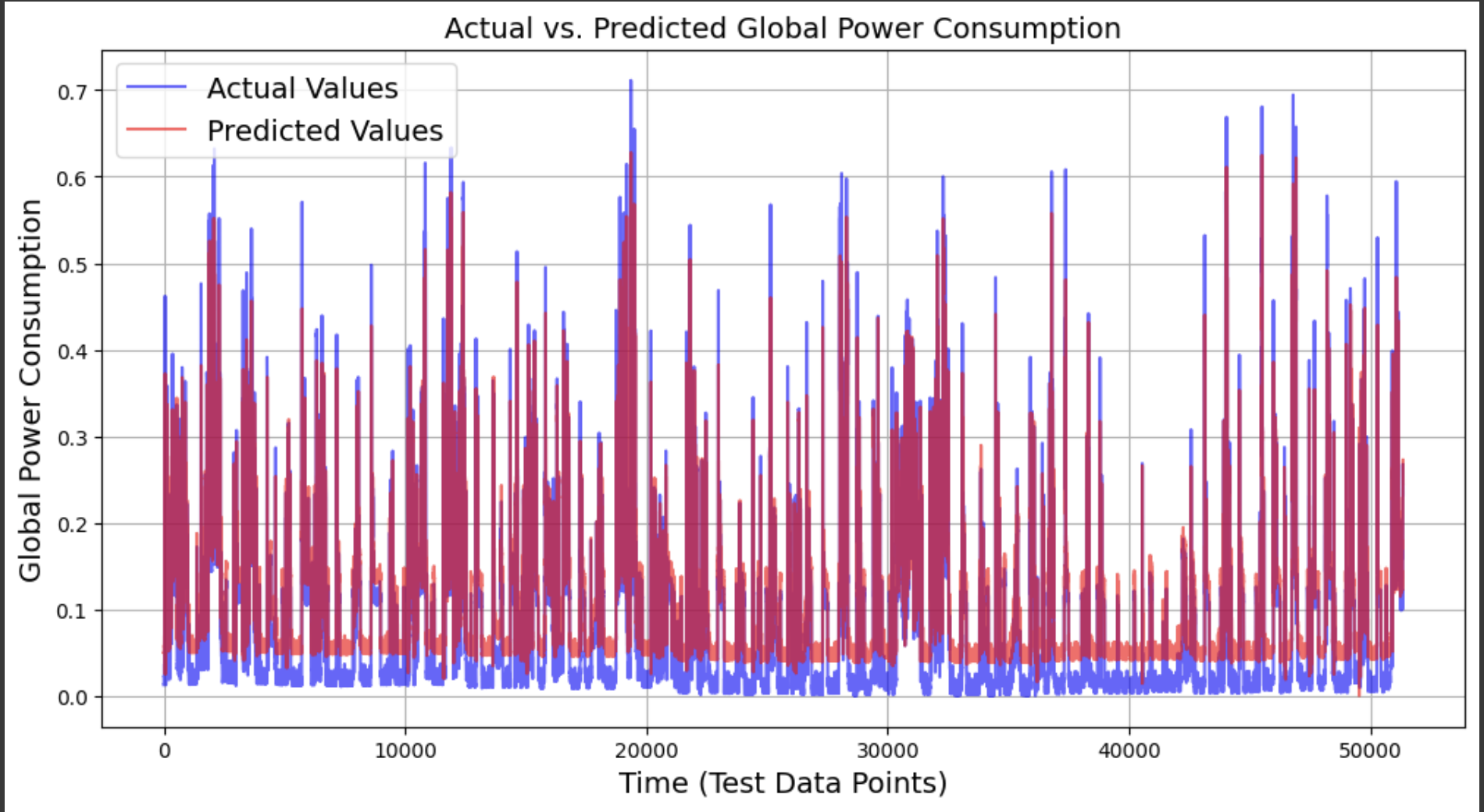}
    \caption{Visualization of Forecasting Through SVM}
    \label{fig:Visualization of Forecasting Through SVM}
\end{figure}

Most importantly the reassurance of overfitting not becoming an issue for short term forecasting due to the choice of this model, reinforced the proposition to implement short term forecasting through the utilization of SVM. There does seem to be a certain discrepancy, with a consistent offset value, between the predicted and actual power consumption. 

\begin{figure}[H]
    \centering
    \includegraphics[width=7cm]{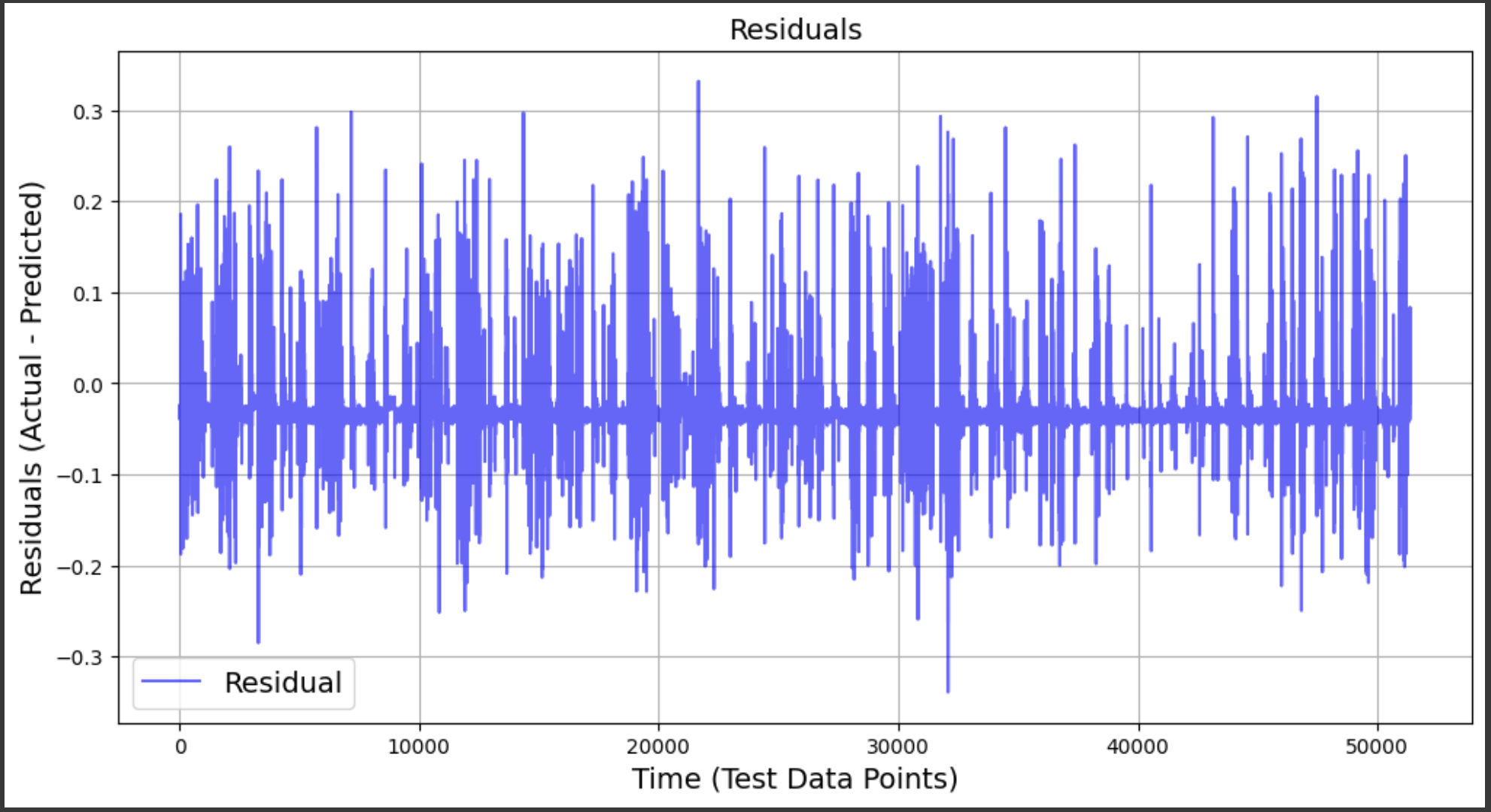}
    \caption{Visualization of Residuals From Short Term Forecasting}
    \label{fig:Visualization of Residuals From Short Term Forecasting}
\end{figure}

While it could pose an issue, similar to which acknowledged to be occurring with the long term forecasting, the acknowledgment of this issue would again allow the IoT system to adapt or configure to handle this consistent offset.
Finally, the anomaly detection was completed on the time series dataset. Through a process of converting the dataset into specific series to capture temporal patterns, and setting certain quantities as bounds to distinguish a value to be an anomaly or not, the ability to implement through an KNN algorithm was presented. 

\begin{figure}[H]
    \centering
    \includegraphics[width=7cm]{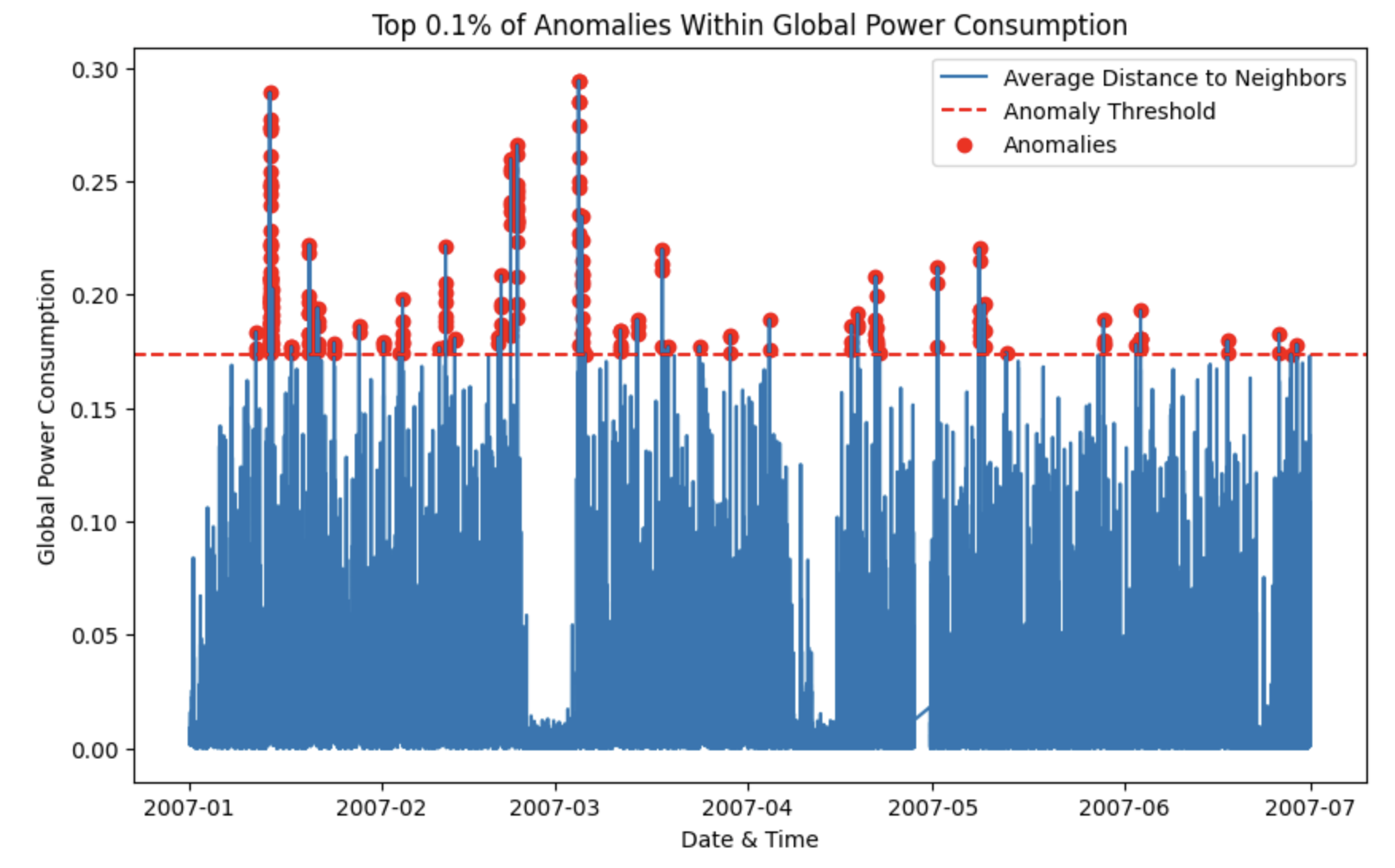}
    \caption{ Top 0.1\% of Anomalies In Power Consumption Records}
    \label{fig:Visualization of Anomalies}
\end{figure}

The visualization of such anomalies as stated before would support certain decisions that would directly affect the household, the city subject to the anomalies, or signal to the power grid of possible issues that require immediate attention. Within the application completed, only 0.1\% of distances measured would be considered to be anomalies. A decision dependent on considering the variability that exists with human behavior and ensuring that anomalies within this context are one that signal the existence of an issue or extreme changes, rather than a reasonable adaptation that an IoT system should integrate. That being acknowledged the presentation of such anomaly detection is possible is the vital focus of the implementation and the anomaly boundary could be varied based on direct preference. 

\section{Conclusion} 
    
The importance of the experimental processes conducted within this study is to reinforce and present that the need and feasibility of a holistic framework. The ability to configure an IoT system that implements long-term forecasting, short-term forecasting, anomaly detection and qualitative consideration would support grid stability and create a proactive approach to energy management. A holistic approach, in addition to supporting households to minimizing power consumption usage, it would allow for greater connective IoT adaptation, in regard to how mass shifts in power consumptions in certain areas would alert of IoT systems of households or neighbors to possibly prepare for such changes, as it may be an effect of an external variable.
While current implementation of such algorithms have been completed in separation, and without the connectivity supportive aspect to it, committing this a certain process to build a foundation for the future of IoT systems would be substantial in standardization of development, rather than different parts being focused and not brought together to have a real buildable impact. While the project only theoretically projected the use of Decision Tree Classifiers, this is another aspect vital to the holistic  prediction and adaptation of IoT devices  to the context of the situation. This is an area that could further be evaluated to identify other considerations that would affect power consumption, in addition to the time, date, weather, and other neighboring IoT system behavior. 
Capitalizing on ensuring a unified, predictive, and adaptive IoT energy framework, this research contributes to creating a foundation for scalable IoT systems, that capable for dynamic adaptation energy management,  enhances situational awareness and logistic decisiveness.

\bibliographystyle{IEEEtran}

\vspace{12pt}
\color{red}.

\end{document}